# Unsupervised Graph Deep Learning Reveals Emergent Flood Risk Profile of Urban Areas

Kai Yin[1], Junwei Ma[1*], Ali Mostafavi[1]

[1] Urban Resilience.AI Lab, Zachry Department of Civil and Environmental Engineering, Texas A&M University, College Station, Texas, United States.

* Corresponding author: Junwei Ma, jwma@tamu.edu.

These authors contributed equally: Kai Yin, Junwei Ma.

**Abstract**

Urban flood risk emerges from complex and nonlinear interactions among multiple features related to flood hazard, flood exposure, and social and physical vulnerabilities, along with the complex spatial flood dependence relationships. Existing approaches for characterizing urban flood risk, however, are primarily based on flood plain maps, focusing on a limited number of features, primarily hazard and exposure features, without consideration of feature interactions or the dependence relationships among spatial areas. To address this gap, this study presents an integrated urban flood-risk rating model based on a novel unsupervised graph deep learning model (called *FloodRisk-Net*). *FloodRisk-Net* is capable of capturing spatial dependence among areas and complex and nonlinear interactions among flood hazards and urban features for specifying emergent flood risk. Using data from multiple metropolitan statistical areas (MSAs) in the United States, the model characterizes their flood risk into six distinct city-specific levels. The model is interpretable and enables feature analysis of areas within each flood-risk level, allowing for the identification of the three archetypes shaping the highest flood risk within each MSA. Flood risk is found to be spatially distributed in a hierarchical structure within each MSA, where the core city disproportionately bears the highest flood risk. Multiple cities are found to have high overall flood-risk levels and low spatial inequality, indicating limited options for balancing urban development and flood-risk reduction. Relevant flood-risk reduction strategies are discussed considering ways that the highest flood risk and uneven spatial distribution of flood risk are formed. The model enables characterizing urban flood risk as an emergent property arising from non-linear interactions among flood hazard and urban features, and integration of spatial flood dependence relationships among areas. Accordingly, this study's outcomes provide a novel perspective and better understanding of urban flood-risk profiles by harnessing machine intelligence and urban big data.

**Introduction**

Urban flooding has been the cause of tremendous social, economic, and ecological impacts throughout the past several decades[1–5]. Catastrophic floods in urban areas, such as those caused by Hurricane Katrina in New Orleans in 2005, Superstorm Sandy in New York in 2012, and Hurricane Harvey in Houston in 2017, have resulted in hundreds of casualties, billions of dollars in property damage, and millions of people adversely affected[6–8]. What is worse, populations and assets in flood-prone areas continue to expand due to uncontrolled urban planning[1,9], surface water runoff is increasing as a result of the growth of impervious surfaces under urbanization, and the precipitation is becoming more frequent and intense due to climate change[1,10,11]. These changes have collectively increased the flood risk in urban areas in particular.

The tremendous impacts caused by floods and the increasing magnitude of flood risk in urban areas bring into sharp focus the importance of characterizing flood risk across areas in cities to identify areas with the greatest risk of floods and to analyze the underlying urban characteristics that give rise to high-flood-risk areas. Identification of high-flood-risk areas and understanding of spatial distribution of flood risk in cities are critical inputs for integrated urban design to achieve flood risk reduction.

The existing approaches for rating urban flood risk are incomplete, inaccurate, and unreliable[1,12]. In the United States, Flood Insurance Rate Maps (FIRMs) provided by Federal Emergency Management Agency (FEMA) are usually used to analyze urban flood risk[1]. FIRMs define flood risk by identifying areas with a 1% chance of being inundated by floods in any given year[13]. Stakeholders (e.g., communities, insurance companies, city managers and planners, and policy makers) merely intersect flood exposure component location information with FIRMs to evaluate the locational flood risk. Several studies[14–17] have revealed, however, that FIRMs perform poorly in evaluating locational flood risk. Datasets from past flood events have also proved inaccurate in predicting urban flood risk. For example, about half the homes in Houston that suffered from flooding caused by Hurricane Harvey, which is a flood event that occurs 3000 to 2000 years, were outside the mapped 500-year floodplains[1]. Relying solely on FIRMs to evaluate urban flood risk causes misunderstanding of locational flood risk, which further leads to the failure of flood risk reduction strategies. Furthermore, FIRMs provide hardly any interpretable information for stakeholders to understand underlying determinants for shaping their flood risk, which hinders them from taking protective actions, such as home improvements, insurance plans, or evacuation during an event.

In the field of flood risk modeling, researchers have made progress in flood risk rating by incorporating flood vulnerability, in contrast to the simplistic real-world approach that merely intersects FIRMs with its exposed components. There are two main streams of methods related to flood risk rating: monetary loss-based and flood risk-related-indicators-based. For the first stream of method, researchers[18–21] generally intersect flood hazard maps with asset location information and their vulnerability function (depth-damage function) to calculate monetary loss caused by floods (e.g., the average annual loss) that is further referred to flood risk. This approach, although widely used, considers only risks associated to limited physical assets (e.g., properties)[22]. It actually measures property damage caused by floods without the ability to rate flood risk considering other aspects of physical and social vulnerabilities. In addition to properties, vulnerability of other physical infrastructure and facilities to floods should be captured. Also, multiple studies[1,23–26] have revealed that floods disproportionately impact socially disadvantaged communities due to their limited capacity to prepare for, resist, and recover from flood damage. Hence, a proper urban flood-risk rating approach should also consider various urban characteristics related to flood hazards, exposure, as well as physical and social vulnerability.

Existing methods proposed for capturing the combined effects of flood hazard, exposure, and vulnerability characteristics in calculating flood risk can be summarized by $\sum w_i * x_i$, where $x_i$ represents flood risk-related features. These approaches focus mainly on determination of the optimized weight value ($w_i$) of each feature using methods, such as the analytic hierarchy process, scoring by experts, set pair analysis, and fuzzy comprehensive evaluation[27]. These weight determination methods are mostly based on expert experience and knowledge[28], resulting in a subjective assessment of flood risk[27]. Furthermore, cities in nature are complex systems with features related to the landscape, populations, and infrastructures that nonlinearly interact with each other given rise to different urban phenomena[29]. Flood hazards in cities interact with other urban features (exposed people and infrastructures, the socially vulnerable communities, and the physically vulnerable infrastructures), which makes the urban flood risk an emergent property arising from complex and nonlinear interactions between flood hazards and urban features. This emergent nature of urban flood risk profiles also implies that independent variables interact with each other[30]. Hence, capturing the complex and non-linear interactions among various urban features is essential for mapping the flood-risk profiles. The existing flood risk-related-indicators-based flood risk assessment methods, however, lack the consideration of interactions among independent variables. Also, indicator-based methods, in essence, calculate flood risk in a linear way, which is inadequate to capture the nonlinear relationships among various hazards and urban features and their intersections that combine to form flood risk.

Another major constraint of the existing approaches for urban flood risk rating is limited consideration of spatial flood dependence relationships among areas of cities. Most of these studies simply assume that the flood risk at a location is independent of all other locations without considering spatial interactions. However, as floods generally spread across many regions, flood risk assessment needs to account for the interrelations of floods between regions[34]. Spatial flood dependence is defined by Brunner et al. as "the degree to which flood risk at one location is related to flood risk at other locations" [31]. Flood frequency in one location will be directly impacted by other locations that have significant spatial flood dependence with it[32]. The flood exposure and flood vulnerability in one location will be indirectly impacted by spatially dependent locations. Therefore, consideration of spatial flood dependence is crucial in determining the flood risk of each location[31,33,34]. Lack of consideration of spatial flood dependence among areas would lead to the underestimate or overestimation of the potential losses caused by flood hazards[34,35].

The purpose of flood risk rating across areas of cities is to identify high risk areas and also enable evaluation of the extent to which urban features related to urban development and growth patterns could shape the flood risk profiles of cities. However, the existing approaches to flood risk rating in cities have major limitations in these aspects. Most of the flood maps are not updated regularly as urban features evolve. An alternative approach would allow updating of flood risk rating of urban areas as the underlying features change over time. Moreover, the existing methods provide a one-size-fits-all rating for the extent of flood risk in cities. The flood risk extent of areas in a city should be determined relative to other areas in the same city. This limitation can be addressed using city-specific rating of flood-risk levels in which a high-risk area in one city would signify a different extent of flood risk than a high-risk area in another city. For example, a level-6 area in city-A does not necessarily have a higher risk than a level-5 area in city-B. The flood-risk levels should be determined based on the specific characteristics and data of each city, making them unique to the local context. City-specific flood-risk level rating enables a better understanding of the flood risk profile across areas of a city.

Recognizing the importance of specifying and characterizing emergent flood risk across urban areas and considering limitations of the existing methods in providing updated flood risk rating, we propose an integrated three-layer unsupervised urban flood risk rating model (called *FloodRisk-Net*) to characterize emergent flood risk in cities using urban big data and machine intelligence. *FloodRisk-Net* is capable of discovering spatial flood-dependence structure, incorporating it into flood-risk rating, and capturing complex and nonlinear interactions among individual flood risk-related features, including flood hazard, exposure, and social and physical vulnerabilities. The five most populous metropolitan statistical areas (MSAs) in the United States are selected as study regions: New York, Los Angeles, Chicago, Dallas, and Houston. The model rates the flood-risk level of grid cells created within each MSA. These five most populous MSAs are chosen considering floods hazards in densely populated areas would cause more serious socioeconomic impacts than other cities. The method we propose can also be generalizable to different MSAs of various population sizes.

The main contribution of this study is mining heterogenous feature information related to urban development and flood hazards and integrating spatial flood dependence relationships using unsupervised graph neural network to reveal emergent property flood risk in urban areas. The model and findings will be particularly valuable to various academic disciplines and diverse stakeholder practitioners: (1) *FloodRisk-Net* provides a novel method for urban scientists and flood researchers to quantify and analyze the flood risk profile of cities based on heterogenous urban development and flood hazard features, and their non-linear interactions; (2) discovering the hierarchical and uneven spatial distribution patterns of flood risk in urban areas informs geo-spatial scientists and urban planners about the spatial autocorrelation and spatial inequality in flood-risk levels and inform integrated urban design strategies in cities; (3) the evaluation of characteristics in high flood-risk-level clusters enables city planners and flood risk managers to develop effective flood risk reduction strategies for balancing urban development and flood-risk reduction.

## Results

**Three-layer unsupervised deep learning model for urban flood risk rating**.

The main tasks associated with urban flood risk rating through machine intelligence and urban big data are: (1) modeling spatial flood dependence relationships and integrating these relationships into flood risk ratings; (2) capturing complex and nonlinear interactions among flood risk-related features; (3) addressing the unavailability of ground truthing labels for flood-risk levels during model training.

To address these tasks, a three-layer unsupervised flood-risk rating model, *FloodRisk-Net* is developed and validated (Fig. 1; see Methods for details). In the first layer, spatial dependence relationship learner, spatial flood dependence relationships among different locations (grid cells) are modeled in a network data structure. The node in the spatial dependence graph is each grid cell, and the weight of the edge connecting two nodes is the spatial dependence strength between them. An unsupervised graph structure learning (GSL) approach is implemented to learn the optimal spatial flood dependence network. The adjacency matrix is modeled by a parameterized graph learner based on the pairwise similarity of node embeddings, considering that spatial flood dependence measures “the degree to which flood risk at one location are related to flood risks at other locations”[31]. Parameters of the graph learner are learned using a contrastive learning framework with training signals obtained from the data itself.

In the second layer, node representation leaner, a node embedding module composed of a deep neural network (DNN) and a graph convolution network (GCN) is implemented to jointly capture complex and nonlinear interactions among flood risk-related features and to incorporate spatial flood dependence relationships. These relationships are incorporated into flood risk rating by propagating and aggregating node attributes from neighbors on the optimal spatial flood dependence network using GCN. This approach originates from the recognition that flood hazard, exposure, and vulnerability in one location would be impacted by the features of other locations upon which it is spatially dependent (see Methods for details). DNN is adopted to capture complex and nonlinear interactions among flood risk-related features in each grid cell. Node embeddings learned by DNN and GCN are combined to obtain the final informative node embeddings.

**Table 1 Selection of flood risk-related features**

| Flood risk aspects | Related features | Data source |
|---|---|---|
| Hazard | Flood intensity | FEMA flood insurance claim dataset |
| | Flood frequency | |
| Exposure | Population number | US Census Bureau dataset |
| | Building area | National Structure Inventory (NSI) dataset |
| Social vulnerability | Poverty rate | The CDC/ATSDR Social Vulnerability Index (SVI) dataset |
| | Disability rate | |
| | Limited English rate | |
| Physical vulnerability | POI number | SafeGraph |
| | Building age | National Structure Inventory (NSI) dataset |
| | Reciprocal foundation height | |

In the third layer, flood-risk-level determination, grid cells with similar flood risk are clustered into the same group, ensuring grid cells in different groups have as much difference in flood risk as possible by implementing a deep clustering framework on the learned node embeddings. The aggregated flood-risk value of each cluster is calculated by multiplying values of flood hazard, exposure, and vulnerability. These aggregated flood risk values are ranked to obtain the flood-risk level of each cluster.

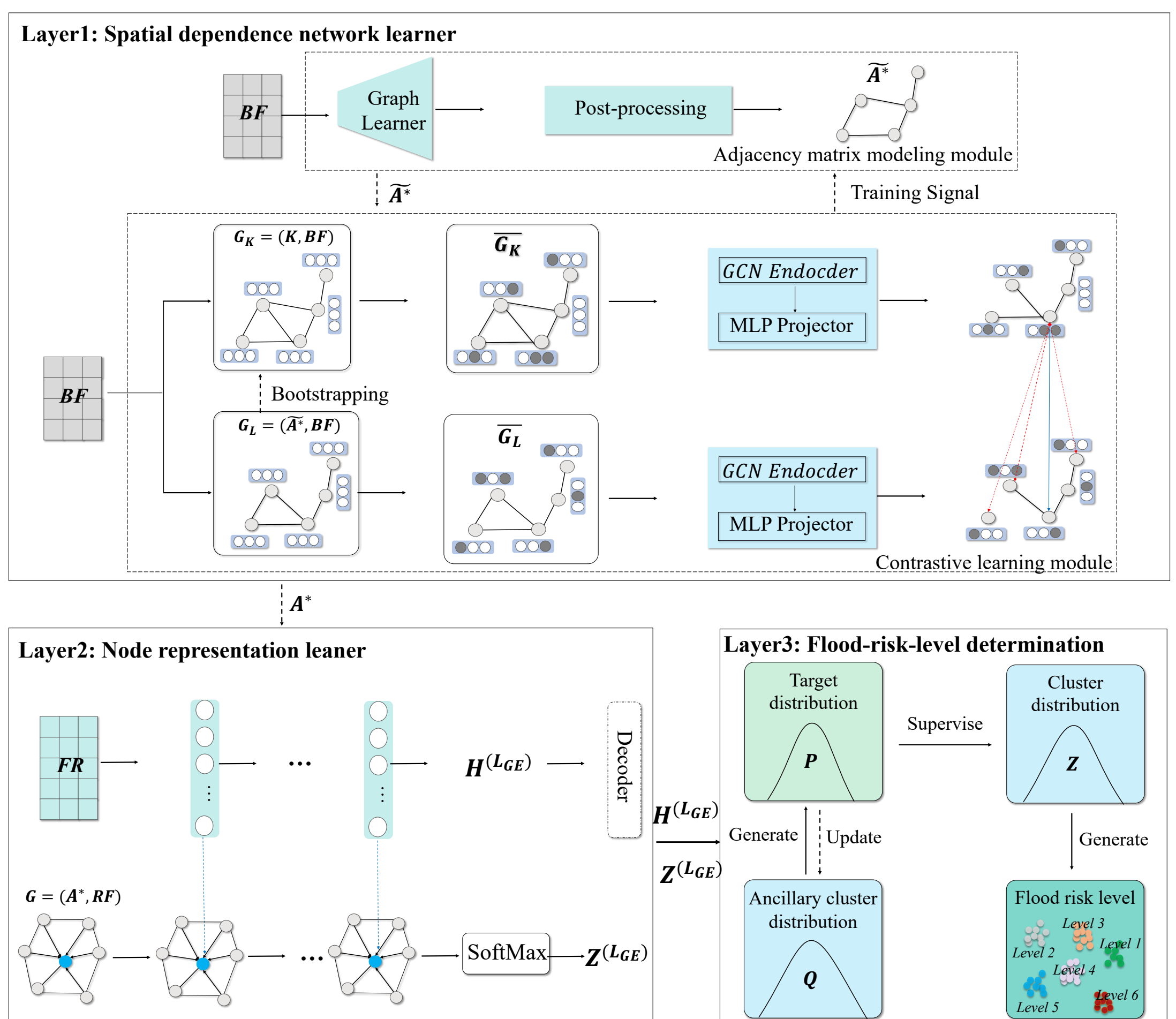

**Fig. 1 | The overall framework of the proposed three-layer urban flood risk-rating model.** Each MSA is divided into grid cells of equal size. Areas in each grid cell are assumed to have the same flood-risk level. The binary flood occurrence matrix ($\boldsymbol{BF}$), measuring whether there is a flood event in grid cell $i$ during week $j$, is taken as the input data to learn the spatial flood dependence network ($\boldsymbol{A}^*$) (See Methods for the construction of $\boldsymbol{BF}$.) The flood risk features matrix ($\boldsymbol{FR}$), which includes flood hazard, flood exposure, and social and physical vulnerabilities information of each grid cell, is constructed (See Methods for the construction of $\boldsymbol{FR}$,). $\boldsymbol{FR}$ is standardized using z-score standardization. The standardized $\boldsymbol{FR}$ is treated as the node attributes, and the optimal adjacency matrix of spatial flood dependence network ($\boldsymbol{A}^*$) is regarded as the graph structure to capture nonlinear interactions among flood risk-related features and to integrate flood spatial dependence relationships. Notations of core parameters used in this model are illustrated in the Supplementary Table S4.

We validated the model by calculating cosine similarity between each grid cell's embedding, as shown in Fig. 2. We find that inner-cluster similarities are significantly larger than between-cluster similarities. Here, inner-cluster similarity means similarity between grid cells in the same cluster, while the between-cluster similarity is the similarity between grid cells in different clusters. Higher inner-cluster similarity indicates that grid cells grouped together share more similar flood-risk characteristics. Lower between-cluster similarity highlights the distinctiveness between different flood-risk levels. These inner-cluster similarities validate that we successfully clustered grid cells with similar flood-risk levels within the same group and ensure the differences between groups as much as possible. This could further validate the spatial dependence learner and the node representation learner as the clustering is conducted based on the node representation, and the node representation is learned based on the spatial dependence.

Between-cluster similarities are found to be larger among similar flood-risk level clusters than those among different flood-risk level clusters. For example, similarities between levels 6 and 5 are significantly larger than those of levels 6 and 1. Areas with similar (but not the same) flood-risk levels would show larger similarities, making similar risk level clusters show larger between-cluster similarities. As we rank data samples in the similarity matrix based on the calculated flood-risk level, this between-cluster similarity could validate the calculation of the flood-risk level for each cluster.

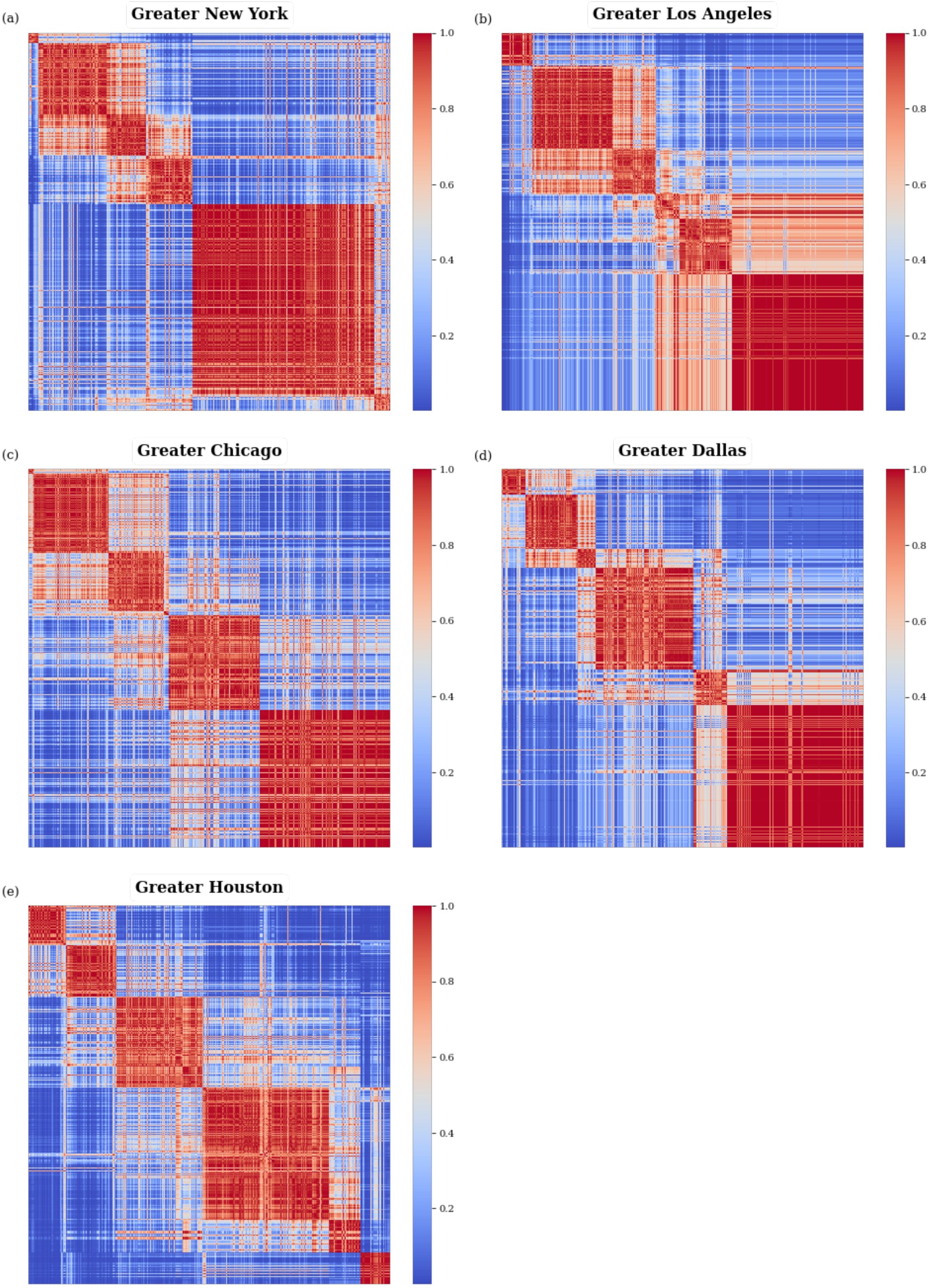

**Fig. 2 | Similarities between each grid cell's embedding for all five MSAs.** Cosine similarity is used to measure the similarity between the embeddings of different grid cells. The similarities are organized by the flood-risk levels assigned to each grid cell. The cosine similarity value ranges from 0 to 1, where 1 indicates identical embeddings. The color intensity in the heatmap indicates the cosine similarity value, with red colors representing high similarity.

**Revealing six flood-risk levels within each MSA under interactions of flood risk-related features and integration of spatial dependence relationships**.

*FloodRisk-Net* yields six flood-risk levels, with level 1 being the lowest flood risk, and level 6. being the greatest. (Fig. 3). Each cluster of risk levels has distinct feature values related to the extent of hazards, exposure, and physical and social vulnerability enabling the interpretation of each flood-risk level (Fig. 4). For example, the greatest flood-risk level areas in Greater Houston are found to have the highest flood hazard, flood exposure, and suffer from the highest social and physical vulnerabilities. Areas with the highest flood risks are situated in or near the Houston city which is located in flat, low-lying regions with high rainfall and clay-based soils[1], characterizing these areas as having the highest flood hazard in this MSA. Furthermore, the largest number of residences and buildings within Greater Houston is in these areas, resulting in the highest flood exposure. Residents in these areas have the highest poverty rate, the highest limited English speaking, and infrastructures in these areas are the oldest with the lowest foundation height and the largest attraction of people (the largest POI number), which make these areas have the highest social and physical vulnerabilities.

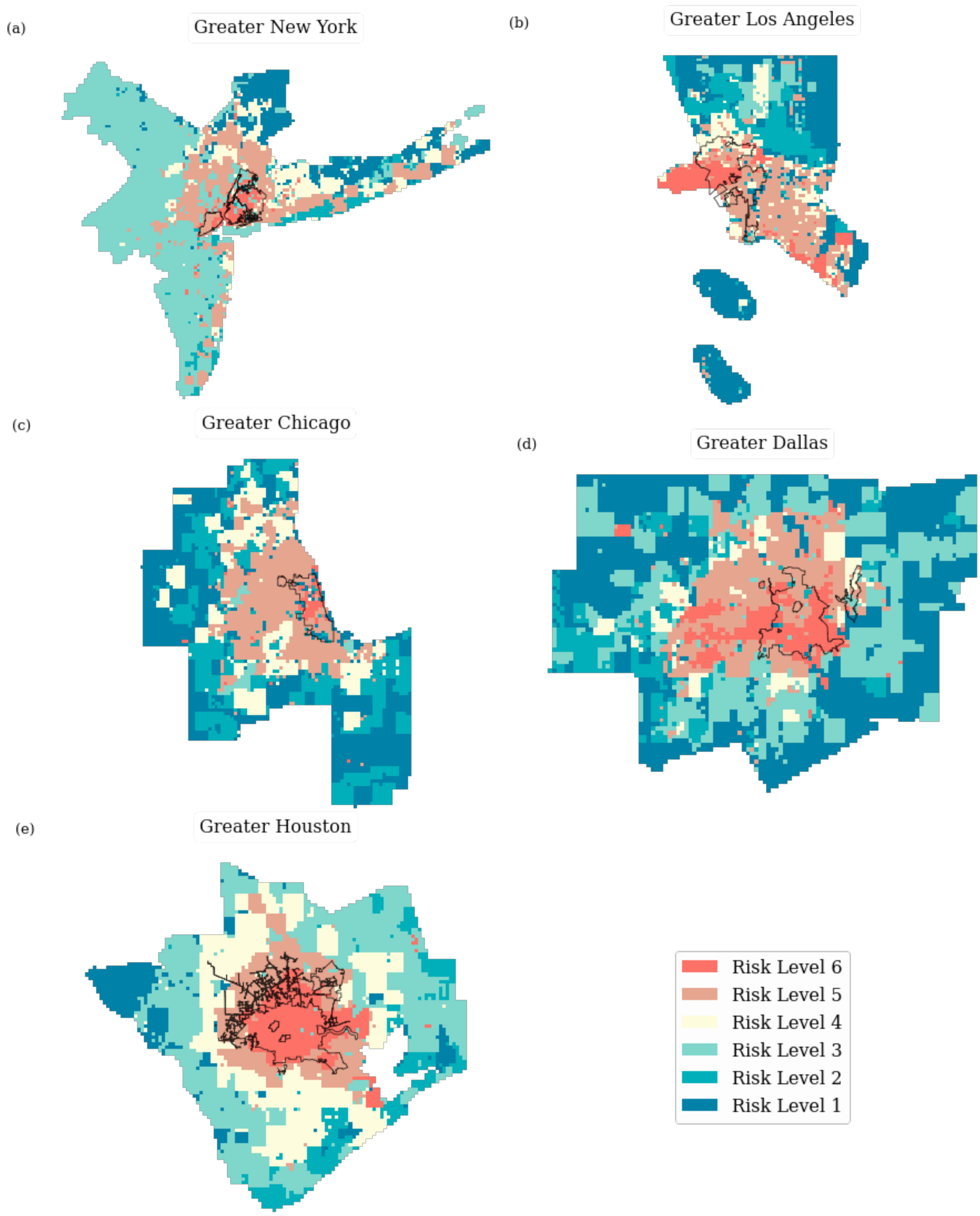

**Fig. 3 | Geographical distribution of flood-risk levels in all five MSAs.** Geographical distribution of flood-risk levels in (a) Greater New York, (b) Greater Los Angeles, (c) Greater Chicago, (d) Greater Dallas, (e) Greater Houston. Core cities (marked as black boundaries) for these five MSAs are (a) New York city, (b) Los Angeles city, (c) Chicago city, (d) Dallas city, and (e) Houston city.

Spatial flood dependence plays a key role in the formation and spatial distribution of the six levels of flood risk. Global Moran's I index is calculated by taking spatial dependence network as spatial weight matrix (see Supplementary Note S7 for its calculation.). The large positive values of Global Moran's I with the small enough $p$-values (Table. 2) quantitively illustrate that grid cells with spatial dependence relationships tend to have similar flood-risk levels. This result shows the presence of spatial autocorrelation of areas with similar flood risk levels, which further suggests that flood risk reduction measures in one area would have possible spillover effects and also reduce the flood risk of neighboring areas. Conversely, development patterns in one area could exacerbate flood-risk level in the neighboring areas.

Table 2. Results of Global Moran's I index in the five MSAs

| MSAs | Greater New York | Greater Los Angeles | Greater Chicago | Greater Dallas | Greater Houston |
|---|---|---|---|---|---|
| Moran's I value | 0.792*** | 0.763*** | 0.830*** | 0.708*** | 0.930*** |

***significant at 0.01 level

**Hierarchical and uneven spatial distribution of flood risk.**

Flood risk is distributed hierarchically from the highest to the lowest level, with the highest flood risk areas being typically encompassed by those with the second-highest risk areas, which are further surrounded by areas with the third-highest, and so forth until the boundary of MSA is reached (Fig. 3). This result indicates that the hierarchical development of cities combined with flood hazards would shape a hierarchical flood risk profile in cities. These results also show that the spatial gradient of flood-risk levels is anisotropic in cities, and the rate of change in flood-risk levels is different in different directions moving from the city core. An anisotropic spatial distribution of flood risk suggests that the risk of flooding varies depending on the direction from the city core. This could be due to city topography, land use, or urban sprawl patterns.

The overlapping of the flood risk map and the core city boundary show that the high flood risk disproportionately falls in the core city (Fig. 3). The flood risk-related features (except for the disability rate) in the core city are larger than those in other areas in the MSA (Fig. 4). This means that the core city in each MSA suffers from greater flood hazards, and more people and infrastructure are exposed to flood hazards with a larger proportion being socially or physically vulnerable. This result signifies the relationship between urban sprawl and development patterns that shape hierarchical structure and the flood risk profile of cities. In a city such as Houston without land use plans and policies, the random development and growth patterns are likely to exacerbate situations in high-flood-risk areas and lead to the cascade of flood-risk levels to the surrounding areas, thus increasing the flood-risk level of areas outside the core city.

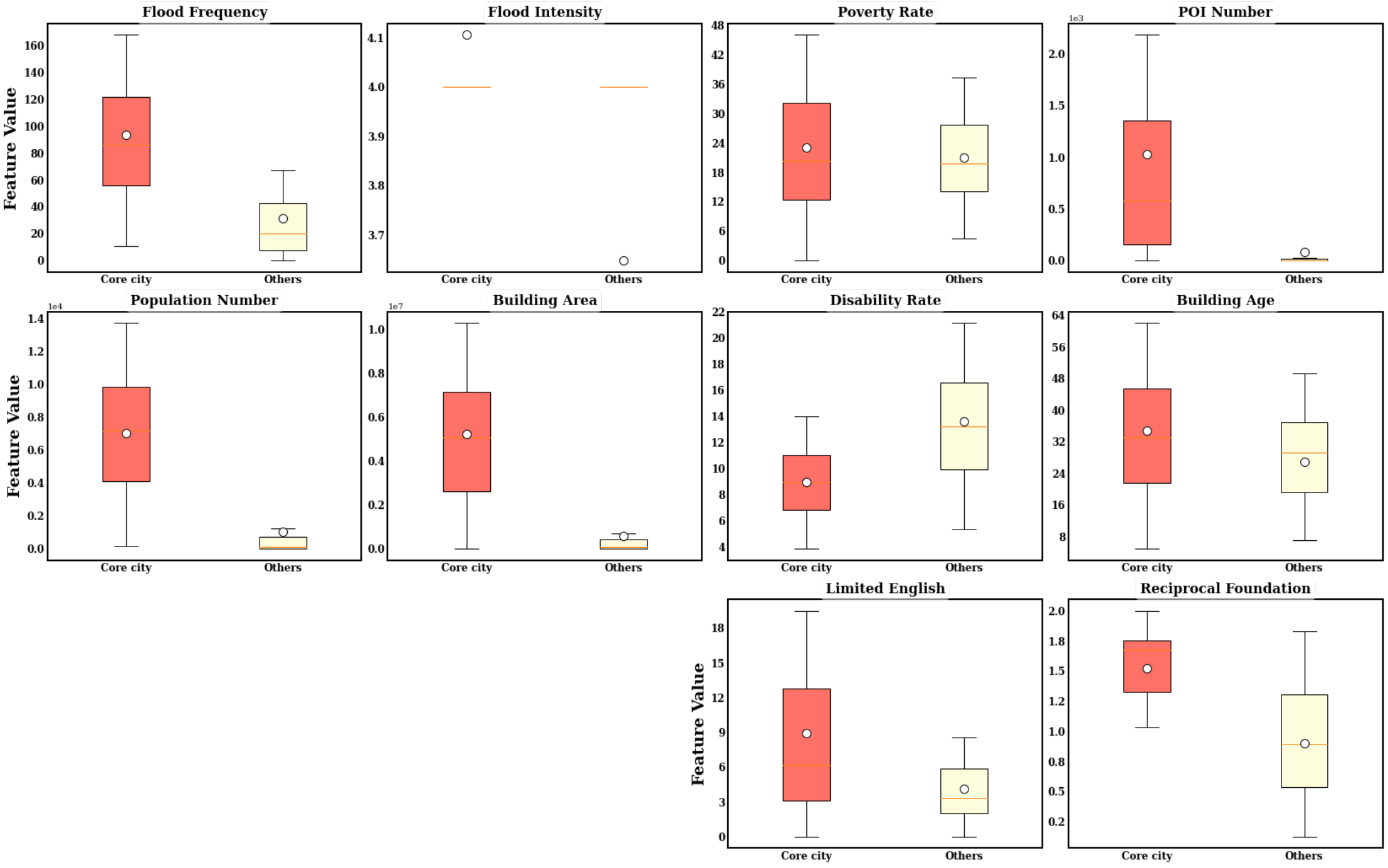

**Fig. 4 | Characteristics of feature distribution in core city and other areas in Greater Houston**. The map for the other four MSAs are shown in Supplementary Fig. 5 to Fig. 8.

In each MSA, cities with a population larger than 25,000 are further selected to examine flood risk distribution. The average flood-risk level and spatial inequality of each selected city are calculated (see Supplementary Note S8). The large negative pearson correlation between average flood risk and spatial inequality of selected cities combined with small enough $p-$ values means that cities with higher average flood-risk levels tend to have smaller spatial inequality (Fig. 5). This indicates that less significant spatial variations of flood risk distribution could be found in cities with higher flood-risk levels. As shown in Supplementary Figure S13, some cities are found to have high flood-risk levels while having low spatial inequality (with the threshold of average flood risk $\geq 5$ and spatial inequality $\leq 0.1$). Such high average flood-risk levels and small spatial inequality should be noted because it indicates that there are few low-risk areas in this city for further urban growth and development. In fact, the spatial inequality of flood-risk levels could be an indicator of the extent to which urban growth and development could be made without exacerbating the overall flood risk of cities. Due to low spatial inequality, further urban development would exacerbate the overall flood risk of the cities since there are limited available low-flood-risk areas. In cities with high risk levels and low spatial inequality, low impact development strategies are essential for balancing urban growth and flood risk reduction plan.

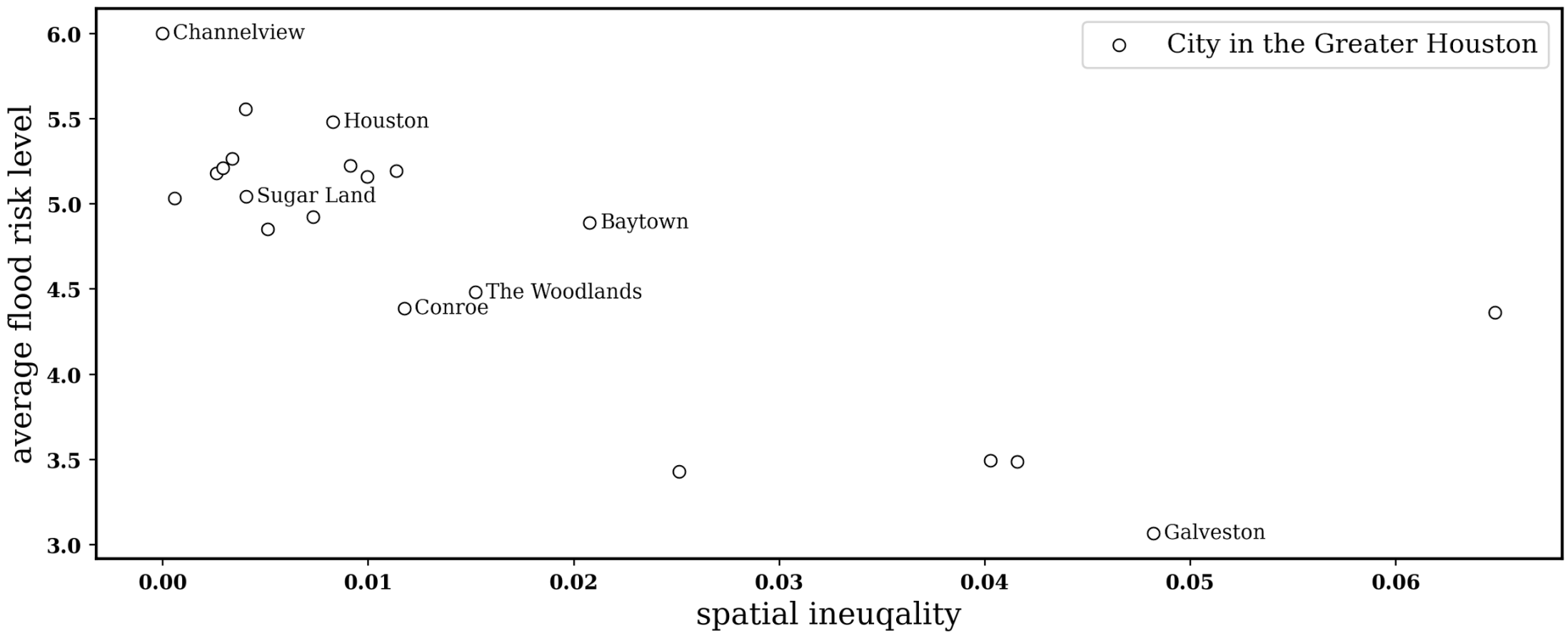

**Fig. 5 | Relation between average flood risk and spatial inequality of cities within Greater Houston.** The other four MSAs are shown in Supplementary Fig. 9 to Fig. 12. The Pearson correlation and $p$-values for average flood risk and spatial inequality of cities in Greater New York, Greater Los Angeles, Greater Chicago, Greater Dallas, and Greater Houston are $(-0.461,\ 6.404 \times 10^{-5})$, $(-0.507,\ 2.861 \times 10^{-8})$, (-0.693, $1.454 \times 10^{-12}$), (-0.653, $6.426 \times 10^{-7}$), and (-0.772, $4.421 \times 10^{-5}$), respectively.

**Three archetypes shaping the highest flood-risk level.**

The underlying determinants shaping the highest flood-risk levels vary across cities, but fall into three archetypes. In Greater Houston, the highest flood-risk level areas fall into the first archetype: high flood hazards, high exposure, and high social and physical vulnerabilities (Fig. 6). In these highest flood risk areas, the largest flood hazard combined with the largest flood exposures cause significant direct impacts (e.g., damage to infrastructure, loss of life). The impact on vulnerable populations and infrastructure is exacerbated due to their limited ability to cope with, resist, and recover from adverse impacts caused by floods[24]. The highest flood risk areas in Greater New York and Dallas are similar to those of Greater Houston (Supplementary Fig. S4).

The highest flood risk areas in Greater Chicago fall into the second archetype (Supplementary Fig. S2), indicating that even if flood events in densely developed areas are neither intense nor frequent, they would still cause great risks due to the large number of people and assets exposed and their limited abilities to cope with, resist, and recover from flood damage. The highest flood risk areas in Greater Los Angeles fall into the third archetype (Supplementary Fig S3), in which the largest flood hazards combined with the medium flood exposure and vulnerabilities resulting in the highest flood risk. These results show the importance of city-specific rating of flood risks since the areas with highest flood-risk levels could have different determinants in different cities. For example, if flood-risk levels in other cities are evaluated based on feature values in high flood-risk levels of Greater Houston, their flood-risk levels would be mischaracterized and underestimated.

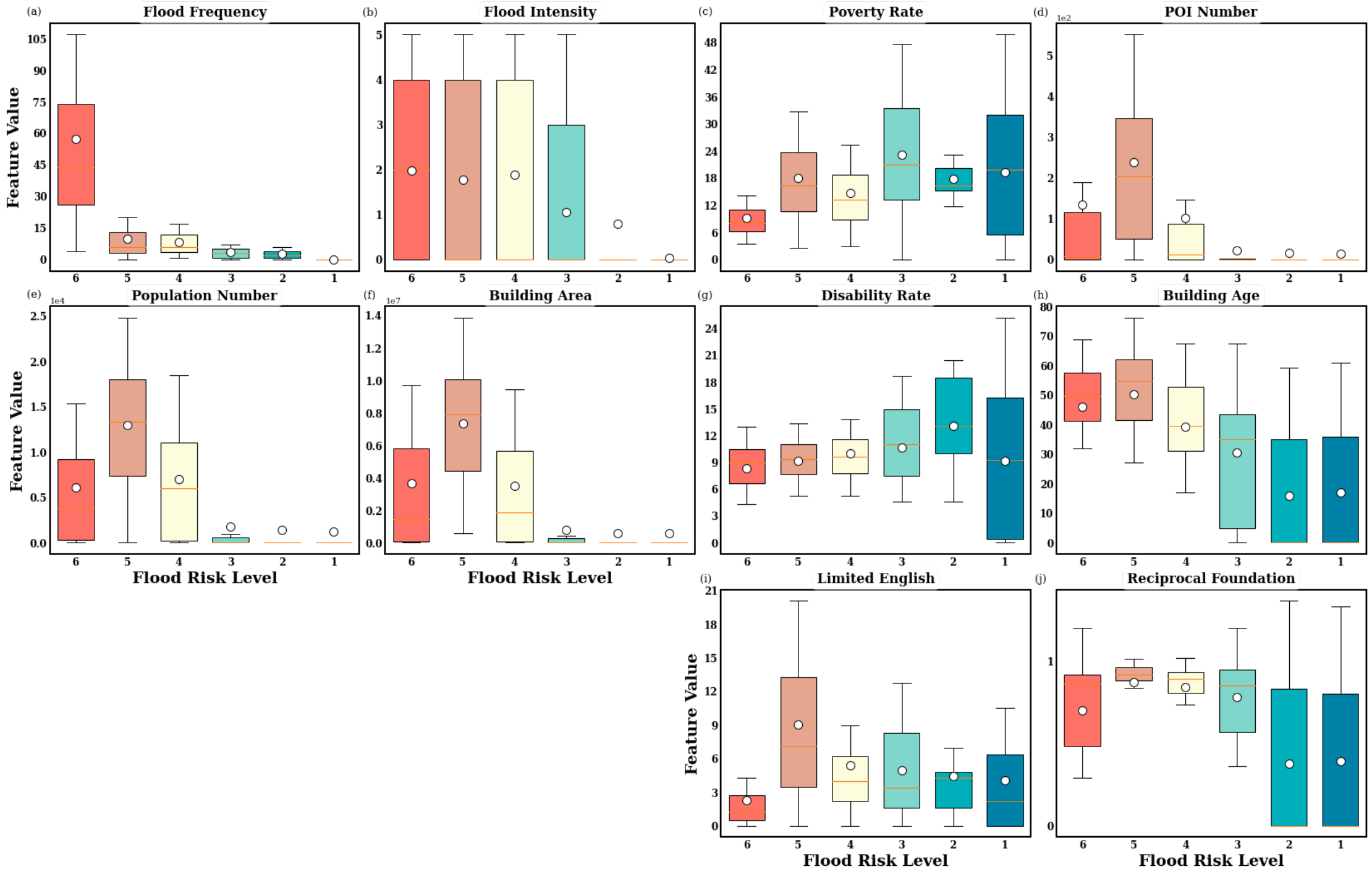

**Fig.6 | Characteristics of clusters in Greater Houston**. Flood risk comprised the features of (a) flood frequency, (b) flood intensity, (c) poverty rate, (d) POI number, (e) population number, (f) building area, (g) disability rate, (h) building age, (i) the percentage of limited English speaking, and (j) reciprocal foundation height. (The maps for the other four MSAs are shown in Supplementary Fig. 1 to Fig. 4.)

Suitable flood risk reduction strategies (summarized in Supplementary Table S3) are examined based on underlying determinants shaping the highest flood-risk levels. Considering the high flood hazards in Level 6 areas, comprehensive stormwater management strategies are necessary to collect, redirect, or hasten the flow of water to reduce flood hazards. Strategies to improve stormwater management include widening ditches, upgrading drainage systems, implementing green infrastructure, and constructing retention ponds. The high flood exposure also calls for the land use planning and policy to limit urban development in high flood-prone areas, or the government's buyout of flood-prone areas. These strategies are important to control the increased exposure in high flood-hazard area. Green infrastructure, such as rain gardens and green roofs, can help to absorb and manage stormwater runoff. In these areas, these strategies can be incorporated into the design of new construction and into the retrofitting of existing buildings. The social vulnerability in these areas is very high with a large percentage in poverty and with limited English proficiency. Financial support should be made available to subsidize flood insurance premiums for poor residents and to assist in the rebuilding of flood-damaged homes or to migrate to less flood-prone areas after floods. Flood information should be translated into languages familiar to the populations to help those with limited English skills become familiar with local flood risks. The physical vulnerability in these areas is also very high, with a large number of POIs, aging buildings, and low foundation heights. Engineering solutions, such as elevating buildings above flood levels, installing flood barriers, and retrofitting buildings, are thus crucial for reinforcing buildings in these areas. Given the large number of POIs in these areas, it is a high priority to identify and protect the most essential infrastructure, such as hospitals, schools, and emergency services, against flood risks. Also, future developments should focus on decentralizing facility distribution to areas with less flood exposure. Urban planning policies that prioritize the allocation of certain essential POIs to low-risk areas are also meaningful for ensuring the continued provision of essential services to the public.

## Discussion and concluding remarks

In this study, we presented an integrated three-layer graph deep learning-based model, *FloodRisk-Net*, for examining the emergent flood risk profiles of cities. *FloodRisk-Net* includes three components: 1) spatial flood dependence relationships within urban areas, 2) node representation learning, and 3) flood-risk-level determination. Using *FloodRisk-Net*, city-specific flood risk ratings are obtained by mining entangled relationships among heterogenous features related to urban development, flood hazards, and the consideration of spatial flood dependence relationships.

The findings provide multiple important contributions to the characterization and analysis of urban flood risk rating. First, departing from the current approaches with binary characterization of flood risk (based on flood-plain maps), the results of this study reveal six distinct city-specific levels of flood risk to evaluate the flood-risk profiles.

Second, the characterization of flood-risk levels obtained from *FloodRisk-Net* capture the complex nonlinear interactions about heterogenous flood hazard and urban development features, as well as spatial dependence of areas. This characterization enables a system-of-systems approach to urban flood risk reduction by applying a holistic and integrative strategy that considers the interdependencies and interactions among various components of an urban environment. It acknowledges that urban flood risk reduction is not just about managing water flows; it also involves land use planning, infrastructure design, and the ramifications of development patterns. By understanding and addressing the complex and nonlinear connections among these components, cities can build more resilient and adaptive systems to mitigate flood risks.

Third, a major burden in planning and implementation of flood risk reduction measures is lack of updated information about flood risk in different areas of cities. In the United States, most floodplain maps are outdated. Significant cost and effort are necessary to produce updated flood hazard maps based on physics-based hydraulic and hydrological models. *FloodRisk-Net*, however, can be implemented conveniently every few years based on updated information about urban development patterns to yield updated flood risk profiles for cities. This dynamic approach ensures that flood risk assessments remain relevant, reflecting changes in urban development and climate conditions.

Fourth, traditional methods provide a one-size-fits-all approach to flood risk assessment, which can lead to mischaracterization of flood risks in different urban contexts. Our model offers a city-specific flood risk rating by clustering grid cells into distinct flood risk levels based on local data. This enables a more accurate representation of flood risk tailored to the unique characteristics of each MSA, thereby improving the applicability of flood risk management strategies across different urban settings.

Fifth, the interpretation of feature values within each flood-risk level, as well as patterns of spatial inequality of flood-risk levels among different urban areas, inform about suitable strategies for balancing urban development and flood risk reduction. Understanding these patterns allows urban planners to design interventions that mitigate risks in the most vulnerable areas, thereby promoting equitable urban development. By incorporating social vulnerability metrics, the methodology ensures that flood risk management strategies are inclusive, addressing the needs of socially disadvantaged communities who are often the most affected by floods. This is critical for achieving social equity in urban development. For example, areas whose high flood-risk levels are due to high flood hazard and high physical and social vulnerability could be targets for flood retreat strategies such as buyouts, zoning policies to restrict or discourage new development in flood-prone areas, and easements. Areas with high flood hazards, high physical vulnerability, and low social vulnerability can be targets for flood infrastructure improvements and flood-proofing requirements (e.g., raising building elevation). Low-impact development strategies are essential for further development in areas with levels 5 and 6 flood risk. While areas with lower flood-risk levels are possible candidates for further development, the spatial autocorrelation of flood risk found in this study suggests that development in areas with lower flood-risk levels would have spillover effect and increase the flood risk of other areas.

*FloodRisk-Net* establishes a new paradigm for understanding emergent flood risk in cities based on urban big data and machine intelligence. It has the potential to be applied to other concepts, such as resilience and reliability through selecting suitable measurement variables and modeling their complex and nonlinear interactions. One unique aspect of *FloodRisk-Net* is the capture of complex interactions among heterogeneous flood risk-related variables, such as flood hazards, exposure, and social and physical vulnerabilities, into latent representations through unsupervised deep graph neural network. Therefore, the method is flexible to change the input variables according to the measured concepts. Machine intelligence enables the use of urban big data to identify the complex interplay between urban features and flood hazards for better understanding and analysis of climate risk and resilience in cities. Developments in the urban big data arena will make the acquisition of large-scale high-resolution urban big data faster and more affordable in the future, thus providing greater potential for our data-driven and machine intelligence-based flood risk rating models by adding additional features based on better and more recent datasets. For example, if data related to future flood exposure under climate change is available or if new flood events occur, the new data could be conveniently added as input features for enhanced characterization of flood risk profiles in cities over time.

There are some important directions for future research. First, the varying degrees of impact of different variables on flood risk estimation could be further considered by investigating the feature importance in flood risk clustering. This would involve analyzing the contribution of each feature to the clustering results, which could provide deeper insights into the factors driving flood risk in different areas. Second, future studies could also explore the application of our model to cities with varying levels of urbanization. This would help to evaluate the model's performance in different urban contexts and enhance its generalizability. Third, while our current methodology maintains a consistent number of clusters across different MSAs to facilitate comparative analysis, we acknowledge the significance of city-specific variations in determining the optimal number of clusters. Considering the unique characteristics and flood risk profiles of different urban areas, future research could benefit from exploring these variations in greater depth. Fourth, the sensitivity analysis of resolution of the grid cell size used to aggregate flood risk-related features could be further evaluated. So that, the impacts of different resolutions of grid cell on the determination of flood risk level could be examined. Fifth, Considering the importance of additional empirical validation of the model's superiority over other clustering models, we recognize the challenge posed by the lack of ground-truth labels for flood risk. This absence is particularly critical when considering flood hazard, flood exposure, social and physical vulnerabilities, and the complex spatial flood dependence relationships. To address this, further experiments could be conducted to investigate the performance of different clustering models in the absence of ground-truth labels. These experiments would help evaluate the robustness and reliability of various models, including *FloodRisk-Net*, under these constraints and enhance our understanding of their comparative effectiveness in real-world applications.

## Methods

### Data processing.

For each MSA, a binary flood occurrence matrix ($\boldsymbol{BF} \in R^{m*d_{BF}}$, $m$ being the number of grid cells in each MSA and $d_{BF}$, the number of weeks during study period) is constructed indicating whether there is a flood event in grid cell $i$ during week $j$[36], using FEMA historical flood claim dataset. Details of the construction of $\boldsymbol{BF}$ are described in Supplementary Notes S1.

Ten flood risk-related features (Table 1), selected from aspects of flood hazard, exposure, and vulnerability following de Moel et al.[37], UNISDR[38], and Wing et al[21].'s work, are adopted to construct the flood risk feature matrix ($\boldsymbol{FR} \in R^{m*10}$). These flood risk-related features are in different scales ranging from census-tract level to highly-detailed geographic coordinates at the resolution of seven decimals. The study area is divided into grid cells of equal size. Flood risk-related features are aggregated into the grid-cell level. The resolution of the grid cell is set to be 2 km ×2 km considering both the computation cost and the need for proper scale of analysis related to urban flood risk profiles. Each column in $\boldsymbol{FR}$ represents a feature and every row records flood risk features for a grid cell. Flood frequency and intensity measure flood hazard using the FEMA historical flood claim dataset[36]. Population and building area, representing the valuated social components (e.g., people and buildings) exposed to flood hazards, are adopted to measure flood exposure using Census Bureau dataset[39] and National Structure Inventory (NSI)[40] dataset, respectively. Flood vulnerability is measured from physical and social aspects. Poverty rate, disability rate, and limited percentage of English-speaking persons are selected to measure social vulnerability using Centers for Disease Control/Agency for Toxic Substances and Disease Registry Social Vulnerability Index (SVI) dataset[41]. POI number, building age, and the foundation height of the building are used to measure physical vulnerability using SafeGraph[42] and National Structure Inventory[40] dataset. Reasons for selecting these features to measure flood risk and details of the construction of $\boldsymbol{FR}$ are described in Supplementary Notes S2.

### Graph structure learner.

We obtain node embedding ($\boldsymbol{E}$) from the input data $\boldsymbol{BF}$ through an embedding network and model the network as the pair-wise similarity between node embeddings as shown in Equation 1. We try to learn optimized graph structures by calculating pair-wise similarities between node embeddings instead of relying on raw node features. Embeddings of node features could provide useful information for learning better graph structures, owing to that parameters of node embedding learning could highlight different dimensions of raw node features[43].

$$\widetilde{\boldsymbol{A}} = \phi(h_w(\boldsymbol{BF})) = \phi(\boldsymbol{E}) \quad (1)$$

where, $h_w(\cdot)$ is the embedding network which adopts neural network to conduct node embedding with the parameter $w$, $\phi(\cdot)$ is cosine similarity function to calculate pairwise similarity, $\boldsymbol{E} \in R^{m*d_{BF}}$ is the node embedding with respect to flood occurrence, and $\widetilde{\boldsymbol{A}} \in R^{m*m}$ is the learned adjacency matrix of the spatial flood dependence graph with unknown parameters.

Multi-Layer Perceptron (MLP)[44] is chosen as the embedding network because it could consider "the correlation and combination of input features", which helps "generate more informative embeddings for similarity metric learning"[45]. A single embedding layer can be written as:

$$\boldsymbol{E}^{(l)} = h_{(w)}^{(l)}(\boldsymbol{E}^{(l-1)}) = \sigma(\boldsymbol{E}^{(l-1)}\boldsymbol{\Omega}^{(l)}) \quad (2)$$

where, $\boldsymbol{E}^{(l)} \in R^{m*d_{BF}}$ is the output matrix of the $l$-th layer of the embedding model, $\boldsymbol{\Omega}^{(l)} \in R^{d_{BF}*d_{BF}}$ is the learnable parameter matrix of the $l$-th layer, $\sigma(\cdot)$ is a non-linear function to make the training more stable. The input in the first layer of $\boldsymbol{E}^{(0)}$ is $\boldsymbol{BF}$, and the output of the last layer $\boldsymbol{E}^{(L_{en})}$ is the obtained node embedding matrix ($L_{en}$ is the number of layers of the graph leaner embedding network). The $L_2$ normalization function is adopted as $\sigma(\cdot)$ to make training more stable, and the activation function of each layer in MLP is set as $ReLU$ function.

A non-negative adjacency matric $\widetilde{\boldsymbol{A}}^{*}$ is further extracted from $\widetilde{\boldsymbol{A}}$ to ensure the value in the matrix ranges from [0, 1], as shown in Equation 3.

$$\widetilde{\boldsymbol{A}}^{*} = p_{act}(\widetilde{\boldsymbol{A}}) = ReLu(\widetilde{\boldsymbol{A}}) \quad (3)$$

**Multi-view graph contrastive learning module.**

A multi-view graph contrastive learning framework (Fig. 1 Layer 1) is developed, with experience from SUBLIME[45], SimCLR[46], and GRACE[47], to acquire supervision signal from data itself for training unknown parameters of graph structure learner. Two graph views are constructed to contrast. The graph structure generated by the graph learner is treated as learned graph view ($\boldsymbol{G}_L = (\widetilde{\boldsymbol{A}}^{*}, \boldsymbol{BF})$), and the graph structure generated by $KNN$ is treated as $KNN$ graph view ($\boldsymbol{G}_K = (\boldsymbol{K}, \boldsymbol{BF})$). ($\boldsymbol{K}$ is $KNN$ graph structure, see Supplementary Notes S3 for details of its construction.) These two views are regarded as being generated from the hidden actual graph structure.

Two frequently used data augmentation methods are adopted (attribute masking and edge perturbation) to corrupt these two views. Details of the data augmentation are described in the Supplementary Notes S4. $\overline{\boldsymbol{BF}}$, $\overline{\boldsymbol{G}_K}$, and $\overline{\boldsymbol{G}_L}$ are the augmented binary flood occurrence matrix, $KNN$ graph view, and learned graph view, respectively. A node-level contrastive learning framework is applied to these two augmented graph views ($\overline{\boldsymbol{G}_K}$ and $\overline{\boldsymbol{G}_L}$) to train unknown parameters in the graph learner. A neural network-based encoder ($f_{\theta}(\cdot)$) is applied for two augmented graph views ($\overline{\boldsymbol{G}_K}$ and $\overline{\boldsymbol{G}_L}$) to obtain representation:

$$\boldsymbol{H}_K = f_{\theta}(\overline{\boldsymbol{G}_K}), \boldsymbol{H}_L = f_{\theta}(\overline{\boldsymbol{G}_L}) \quad (4)$$

where, $\theta$ is the parameter of the encoder; it is the same for the two views. $\boldsymbol{H}_K$ and $\boldsymbol{H}_L \in R^{m*d_1}$ ($d_1$ is the encoded representation dimension) are the encoded node representations for augmented $KNN$ graph and learned graph views, respectively. We choose GCN as encoder network[48].

An MLP-based projector ($g_{\xi}(\cdot)$) is further applied on the encoded node representations to map them to the space where contrastive loss is applied.

$$\boldsymbol{Z}_K = g_{\xi}(\boldsymbol{H}_K), \boldsymbol{Z}_L = g_{\xi}(\boldsymbol{H}_L) \quad (5)$$

where $\xi$ is the parameter of the projector $g_{\xi}(\cdot)$; it is the same for both views. $\boldsymbol{Z}_K$ and $\boldsymbol{Z}_L \in R^{m*d_2}$ ($d_2$ is the projected representation dimension) are the projected node representations for augmented $KNN$ graph and learned graph views, respectively.

A symmetric normalized temperature-scaled cross-entropy loss ($NT - Xent$)[46,49,50] is adopted as the contrastive loss function. $NT - Xent$ aims to maximize the mutual information between $z_{k,i}$ and $z_{l,i}$ of node $v_i$ on two views.

$$L_G = \frac{1}{2n}\sum_{i=1}^{n}[l(z_{k,i}, z_{l,i}) + l(z_{l,i}, z_{k,i})] \quad (6)$$

$$l(z_{k,i}, z_{l,i}) = -log\frac{e^{\phi(z_{k,i}, z_{l,i})/t}}{\sum_{i=1}^{n} 1_{[i \neq j]} e^{\phi(z_{k,i}, z_{l,j})/t}} \quad (7)$$

where $t$ is the temperature parameter, $l(z_{k,i}, z_{l,i})$ is the loss function for a positive sample $(z_{k,i}, z_{l,i})$, and $1_{[i \neq j]} \in \{0,1\}$ is an indicator function that equals to 1 if $i \neq j$. The loss function for the positive sample $(z_{l,i}, z_{k,i})$ is calculated similarly as $l(z_{k,i}, z_{l,i})$.

The final loss $L_G$ is computed as the average value across all positive pairs, and it is minimized to maximize the agreement between two views. The final optimal adjacency matrix of spatial flood dependence network $A^{*}$ is obtained after minimizing $L_G$.

**Bootstrapping.**

To prevent the learned graph view from inheriting error information and over-fitting the $KNN$ graph view, a bootstrapping mechanism[45] is applied to the adjacency matrix of the $KNN$ graph view as follows:

$$K \leftarrow \tau K + (1 - \tau)\widetilde{\boldsymbol{A}}^{*} \quad (8)$$

where $\tau \in [0,1]$ is a coefficient controlling the update of $K$. The adjacency matrix of the $KNN$ graph view will be updated by the adjacency matrix of the learned graph view in a slow-moving step every 10 iterations instead of keeping unchanged.

**Incorporate spatial flood dependence relationships.**
Flood frequency in one grid cell will be directly impacted by other grid cells that have spatial flood dependence with it[33]. Hence, the flood exposure and flood vulnerability in this grid cell will be indirectly impacted by these spatially dependent grid cells.

This role of spatial flood dependence relationship in flood risk rating is incorporated in node representation learning by aggregating node attributes ($\boldsymbol{FR}$) from neighbors on the spatial flood dependence network ($\boldsymbol{A}^*$) using GCN. The representation learned by the $l-th$ layer of GCN, $\boldsymbol{Z}^{(l)}$, can be expressed by the following convolutional operation:

$$\boldsymbol{Z}^{(l)} = ReLu(\widetilde{\boldsymbol{D}}^{-\frac{1}{2}}\overline{\boldsymbol{A}^*}\widetilde{\boldsymbol{D}}^{-\frac{1}{2}}\boldsymbol{Z}^{(l-1)}\boldsymbol{W}^{(l-1)}) \quad (9)$$

where, $\overline{\boldsymbol{A}^*} = \boldsymbol{A}^* + \boldsymbol{I}$ and $\widetilde{\boldsymbol{D}_{ii}} = \sum_j \overline{\boldsymbol{A}^*}_{ij}$. $\boldsymbol{I}$ is the identity diagonal matrix of the adjacency matrix $\boldsymbol{A}^*$. $\boldsymbol{W} \in R^{d_{in}*d_{out}}$ is the learnable parameter matrix. $\boldsymbol{Z}^{(0)}$ is $\boldsymbol{FR}$ and $\boldsymbol{Z}^{(L_{GE})}$ is the output data.

**Capture complex interactions among features.**
An autoencoder framework is adopted, where DNN is adopted as the encoding network to capture complex and nonlinear interactions among individual flood risk-related features.

In the encoder part, node representation learned by the $l-th$ layer of DNN, $H^{(l)}$, can be expressed as follows:

$$\boldsymbol{H}^{(l)} = ReLu(\boldsymbol{W}_e^{(l)}\boldsymbol{H}^{(l-1)} + \boldsymbol{b}_e^{(l)}) \quad (10)$$

where $\boldsymbol{W}_e^{(l)}$ and $\boldsymbol{b}_e^{(l)}$ are the weight matrix and bias of the $l$-th layer in the encoding part. $\boldsymbol{H}^{(0)}$ is the input data and $\boldsymbol{H}^{(L_{GE})}$ is the output data.

The decoder aims to reconstruct the input data through several fully connected layers as follows:

$$\boldsymbol{H_d}^{(l)} = ReLu(\boldsymbol{W}_d^{(l)}\boldsymbol{H}^{(l-1)} + \boldsymbol{b}_d^{(l)}) \quad (11)$$

where $\boldsymbol{W}_d^{(l)}$ and $\boldsymbol{b}_d^{(l)}$ are the weight matrix and bias of the $l$-th layer in the encoding part, respectively.

The autoencoder aims to reconstruct the input data without missing any information, which leads to the following reconstruction loss function:

$$L_{res} = \frac{1}{2N}\sum_{i=1}^{N} || h_i - \hat{h}_i||_2^2 \quad (12)$$

where $h_i$ is the $i$-th row data of input data $\boldsymbol{FR}$ and $\hat{h}$ is the $i$-th row data of $\boldsymbol{H_d}^{(L)}$.

**Node representation integration.**
Node representation $\boldsymbol{H}^{(l-1)}$ learned from DNN preserves the latent information of complex and nonlinear interactions among individual risk-related features for each grid cell. Node representation $\boldsymbol{Z}^{(l-1)}$ incorporates spatial flood dependence relationships among different grid cells. We thus try to combine these two node representations:

$$\widetilde{\boldsymbol{Z}}^{(l-1)} = (1-\epsilon)\boldsymbol{H}^{(l-1)} + \epsilon\boldsymbol{Z}^{(l-1)} \quad (13)$$

where $\epsilon$ is the balance coefficient, we set $\epsilon$ as 0.5 to obtain an informative representation[51].

Then, $\widetilde{\boldsymbol{Z}}^{(l-1)}$ is adopted as the input of the $l$-layer in GCN to learn the representation $\boldsymbol{Z}^{(l)}$:

$$\boldsymbol{Z}^{(l)} = ReLu(\widetilde{\boldsymbol{D}}^{-\frac{1}{2}}\overline{\boldsymbol{A}^*}\widetilde{\boldsymbol{D}}^{-\frac{1}{2}}\widetilde{\boldsymbol{Z}}^{(l-1)}\boldsymbol{W}^{(l-1)}) \quad (14)$$

The representation $\boldsymbol{H}^{(l-1)}$ learned from each DNN layer is integrated with the representation learned from the corresponding GCN layer using Equation 13, as shown in Layer 2 of Fig 1. $\boldsymbol{Z}^{(L_{GE})}$ is the output node representation, where $L_{GE}$ is the layer number of GCN or encoding network of DNN.

**Soft Assignment.**
Grid cells with similar flood risk are clustered into the same group based on node embeddings $\boldsymbol{Z}^{(L_{GE})}$, as $\boldsymbol{Z}^{(L_{GE})}$ capture both nonlinear interactions among flood risk-related features and integrate spatial flood dependence relationships. A SoftMax function is implemented on the output of the GCN module following the work of Bo et al[51]:

$$Z = softmax\,(\widetilde{\boldsymbol{D}}^{-\frac{1}{2}}\overline{\boldsymbol{A}^{*}}\widetilde{\boldsymbol{D}}^{-\frac{1}{2}}\boldsymbol{Z}^{(L_{GE})}\boldsymbol{W}^{(L_{GE})}) \tag{15}$$

The most likely cluster $k$ that instance $i$ belongs ($s_i$) to is.

$$s_i = \underset{k}{\operatorname{argmax}}(z_{ik}) \tag{16}$$

where $z_{ik} \in Z$ is the probability that sample $i$ belongs to the cluster $k$ and $s_i$ is the clustering that grid cell $i$ belongs to.

**"Self-training" strategy.**

A dual self-training strategy[51] is adopted to train the clustering module. The KL-divergence loss between cluster assignment distribution $z_{ik}$ and the target distribution $p_{ik}$ is minimized to obtain the optimal cluster assignment[52] as follows:

$$L_{clu} = KL(P||Z) = \sum_{i}\sum_{k} p_{ik} log \frac{p_{ik}}{z_{ik}} \tag{17}$$

where $P$ is the target distribution. The target distribution $P$ works as the "ground-truth labels" in the training process[53]. Reasons for its eligibility for being selected as the training signal are summarized in Supplementary Notes S5 and it is obtained as follows:

1. The student's $t$-distribution is firstly used to calculate the ancillary cluster assignment distribution $Q$ based on node embedding $\boldsymbol{H}^{(L_{GE})}$, following Maaten and Hinton's[54] work:

$$q_{ik} = \frac{(1 + ||h_i - u_k||_2^2/2)^{\frac{-(\alpha+1)}{2}}}{\sum_j^K (1 + ||h_i - u_j||_2^2/2)^{\frac{-(\alpha+1)}{2}}} \tag{18}$$

   where, $q_{ik}$ is the ancillary cluster assignment distribution that measures the probability of sample $i$ assigned to the cluster $k$, $h_i$ is the $i$-th row of node embedding $\boldsymbol{H}^{(L_{GE})}$, $u_k$ is the cluster center which is initialized by $K$-means on node embedding $\boldsymbol{H}^{(L_{GE})}$ and $\alpha$ is the freedom degree of the student's $t$-distribution, and we set $\alpha = 1$ following Xie et al.'s work[52].

2. The target distribution $P$ is set as the variant of the ancillary cluster assignment $Q$ following Xie et al.'s work[52]. It is computed by firstly raising $q_i$ to the second power and then normalizing it based on the frequency as the following:

$$p_{ik} = \frac{{q_{ik}}^2/f_k}{\sum_j ({q_{ij}}^2/f_j)} \tag{19}$$

   where, $f_k = \sum_i q_{ik}$ is the soft cluster frequencies and can be viewed as "the sum of the probability that instance $i$ belongs to the $k$-th cluster"[55].

To obtain the initial target distribution $P$, the DNN module is first pre-trained to obtain an initial node embedding $\boldsymbol{H}^{(L_{GE})}$ by minimizing $L_{res}$, as shown in Equation 12. $K$-means clustering is conducted once to obtain the initial clustering centers $u$ on the node embedding $\boldsymbol{H}^{(L_{GE})}$ before training the whole clustering model. The initial target distribution $P$ is calculated using Equation 19 after ancillary distribution $Q$ is calculated using Equation 18.

Then, in the main training process, clustering centers $u$ and node embedding $h$ are updated based on gradients of $L_{ta}$ with respect to $u$ and $h$, which could help node embedding closer to the clustering center [51–53]:

$$L_{ta} = KL(P||Q) = \sum_{i}\sum_{k} p_{ik} log \frac{p_{ik}}{q_{ik}} \tag{20}$$

where, $L_{ta}$ is the KL-divergence loss between ancillary cluster assignment distribution $q_{ik}$ and the target distribution $p_{ik}$. Distribution $Q$ is updated in each iteration using Equation 18 with the updated $h$ and $u$. Target distribution $P$ is updated every 5 iterations with the updated $Q$ to keep training process stable using Equation 19 following Wang et al.'s work[53].

The overall loss function is defined as follows:

$$L = L_{res} + \alpha L_{clu} + \beta L_{ta} \tag{21}$$

where $\alpha$ and $\beta$ are coefficients to balance the training of the GCN module and the update of $\boldsymbol{H}^{(L_{GE})}$ between reconstruction loss, respectively. Clustering results are obtained using Equation 16 after training until the maximum epochs.

**Flood-risk-level calculation.**

Grid cells in the same group are assumed to have the same flood-risk level as the clustering module has clustered grid cells with similar flood risk into the same group and ensures that grid cells in different clusters are different from each other as much as possible. The flood-risk level of each cluster is calculated as the following steps:

1. Following Mojaddadi et al.[56], Siam et al.[57], and Xu et al.[58]'s work, flood risk-related features are normalized to reduce the impact of the difference of unit using Min-Max scaler as the following equation:

$$x' = \frac{x - x_{min}}{x_{max} - x_{min}} \tag{22}$$

   where, $x'$ is the scaled value which is in the range of $[0,1]$.

2. We calculate each flood risk-related feature's mean values for each cluster:

$$\overline{\boldsymbol{F}_{\imath ku}} = \frac{\sum_{l=1}^{n} F_{iku_l}}{n} \tag{23}$$

where, $\overline{\boldsymbol{F}_{\imath ku}}$ is the average value for feature $i$ in cluster $k$ within MSA $u$, and $F_{iku_l}$ is the $l$-th value of feature $i$ in cluster $k$ within MSA $u$.

3. We calculate the value of flood hazard, flood exposure, and flood vulnerability of each cluster:

$$\boldsymbol{FH}_{ku} = \overline{\boldsymbol{FF}_{ku}} + \overline{\boldsymbol{FI}_{ku}} \tag{24}$$

$$\boldsymbol{FE}_{ku} = \overline{\boldsymbol{PN}_{ku}} + \overline{\boldsymbol{BA}_{ku}} \tag{25}$$

$$\begin{aligned} \boldsymbol{FV}_{ku} &= \overline{\boldsymbol{SN}_{ku}} + \overline{\boldsymbol{PN}_{ku}} \\ &= \overline{\boldsymbol{PoR}_{ku}} + \overline{\boldsymbol{D\imath R}_{ku}} + \overline{\boldsymbol{LER}_{ku}} \\ &\quad + \overline{\boldsymbol{POI}} + \overline{\boldsymbol{FdH}} + \overline{\boldsymbol{BAg}} \end{aligned} \tag{26}$$

where, $\boldsymbol{FH}_{ku}$, $\boldsymbol{FE}_{\boldsymbol{ku}}$, $\boldsymbol{FV}_{ku}$, $\overline{\boldsymbol{FF}_{\boldsymbol{ku}}}$, $\overline{\boldsymbol{FI}_{ku}}$, $\overline{\boldsymbol{PN}_{ku}}$, $\overline{\boldsymbol{BA}_{ku}}$, $\overline{\boldsymbol{SN}_{ku}}$, and $\overline{\boldsymbol{PN}_{\boldsymbol{ku}}}$ are the flood hazard, flood exposure, flood vulnerability, average flood frequency, average flood intensity, average population number, average building area, social vulnerability, physical vulnerability value of cluster $k$ in MSA $u$, respectively.

4. Following the work of Xu et al.[58], the aggregated flood risk value $\boldsymbol{FR}_{ku}$ of each cluster is calculated by multiplying values of flood hazard, exposure and vulnerability of each cluster:

$$\boldsymbol{FR}_{ku} = \boldsymbol{FH}_{ku} * \boldsymbol{FE}_{ku} * \boldsymbol{FV}_{ku} \tag{27}$$

5. Finally, we calculate the flood-risk level value $\boldsymbol{FL}_{ku}$ by ranking the value of $\boldsymbol{FR}_{ku}$ in an ascending way:

$$\boldsymbol{FL}_{ku} = Sorted\ (\boldsymbol{FR}_{ku}) \tag{28}$$

**Model training.**

Hyperparameters tuning, the algorithm of the urban flood risk rating model, and the determination of the optimal shared clustering number for all five MSAs are described in the Supplementary Notes S6. Our methodology sets the clustering number to be the same across different MSAs to enable a robust comparative analysis of flood risk spatial distribution. By maintaining a consistent number of clusters (i.e., flood risk levels) across all MSAs, we can effectively compare the spatial distribution of flood risk. This uniformity allows us to identify and analyze patterns and discrepancies in flood risk profiles among different cities, providing valuable insights into urban flood risk management on a broader scale. As detailed in Supplementary Note S6, we determine the optimal shared clustering number by calculating the average Calinski and Harabasz (C&H) score across all MSAs. This approach ensures that the chosen clustering number is not only optimal for individual cities but also facilitates meaningful comparisons. Our analysis indicated that six clusters (flood risk levels) provided the highest average C&H score, thus balancing both city-specific optimization and comparative consistency. A consistent clustering framework could simplify the interpretation of flood risk levels for stakeholders, including city planners, policymakers, and emergency management officials. It allows for standardized flood risk mitigation strategies and policies that can be adapted and compared across different cities, fostering collaboration and shared learning.

**Data Availability**
The dataset used in this paper are publicly accessible and cited in this paper.

**Code Availability**
The code that supports the findings of this study is available from the corresponding author upon request.

**Acknowledgements**
This work was supported by the National Science Foundation under CRISP 2.0 Type 2 No. 1832662 grant. Any opinions, findings, conclusions, or recommendations expressed in this research are those of the authors and do not necessarily reflect the view of the funding agencies.

**Author contributions**
**Kai Yin**: Conceptualization, Methodology, Software, Formal analysis, Investigation, Writing – original draft, Visualization. **Junwei Ma**: Conceptualization, Formal analysis, Writing – original draft, Writing – review & editing. **Ali Mostafavi**: Conceptualization, Methodology, Writing—Reviewing and Editing, Supervision, Funding acquisition.

**Competing interests**

The authors declare no competing interests.

**Additional information**

Supplementary material associated with this article can be found in the attached document.